\documentclass{article}

\pdfoutput=1 

\PassOptionsToPackage{numbers,sort}{natbib}
\usepackage[final]{neurips_perls_xt_2021}

\RequirePackage{amsthm,thmtools}

\theoremstyle{definition}

\newtheorem{safety property}[safetyprop]{Safety Property}

\usepackage{amsmath}
\usepackage{amsfonts}
\usepackage{url}

\usepackage[font=it,labelfont=bf,margin=10pt]{caption}

\usepackage[table]{xcolor}
\definecolor{darkgreen}{RGB}{0,64,0}
\definecolor{lightgreen}{RGB}{210,255,210}
\definecolor{lightblue}{RGB}{240,255,255}

\RequirePackage{tikz}
\usetikzlibrary{arrows.meta}
\tikzset{>={latex}}
\usetikzlibrary{arrows.meta, angles, quotes, shapes.misc}
\usetikzlibrary{calc}
\usetikzlibrary{positioning}
\usetikzlibrary{shapes,shapes.geometric}
\usetikzlibrary{arrows}
\usetikzlibrary{fit}
\usetikzlibrary{bayesnet}

\usepackage[linkcolor=black,bookmarksopen,bookmarksopenlevel=2,breaklinks=true]{hyperref}

\begin{document}
 
\title{Demanding and Designing\\ Aligned Cognitive Architectures}
\author{{\bf Koen Holtman}\\[.5ex]
{\normalsize Eindhoven, The Netherlands}\\
{\normalsize \tt Koen.Holtman@ieee.org}}
\date{XX 2021}
\maketitle
\begin{abstract}
With AI systems becoming more powerful and pervasive, there is
increasing debate about keeping their actions aligned with the broader
goals and needs of humanity.  This multi-disciplinary and
multi-stakeholder debate must resolve many issues, here we examine
three of them.
The first issue is to clarify what demands stakeholders might usefully make
on the designers of AI systems, useful because the technology exists
to implement them.
We make this technical topic more accessible by
using the framing of cognitive architectures.
The second issue is to move beyond an analytical
framing that treats useful intelligence as being reward
maximization only.  To support this move, we define several AI
cognitive architectures that combine reward maximization with other
technical elements designed to improve alignment.
The third issue is how stakeholders should calibrate their
interactions with modern machine learning researchers.  We consider
how current fashions in machine learning create a narrative pull that
participants in technical and policy discussions should be aware of,
so that they can compensate for it.
We identify several technically tractable but currently unfashionable
options for improving AI alignment.
\end{abstract}


\section{Introduction}

With AI systems becoming more powerful and pervasive, there is
increasing debate about keeping them aligned with the broader 
goals and needs of humanity.  Good and general introductions to this debate
are Christian's recent book {\it The Alignment Problem}
\cite{christian2020alignment} and Russell's {\it Human Compatible}
\cite{russell2019human}.

Multi-stakeholder discussions about AI alignment can encounter many
barriers to progress, barriers that prevent moving the discussion
forward towards making specific actionable demands and building a
political consensus around them.  In this paper, we aim to lower three
of these barriers: stakeholder uncertainty about what is technically
possible, a too narrow focus on reward maximization, and the narrative
pull of current fashions in machine learning.

In modern AI research, thought experiments like the Turing test have
fallen out of fashion as way to define intelligence.  Instead,
intelligence is usually equated with the ability of an autonomous
system to pick actions which will efficiently maximize a reward
metric.  General-purpose intelligence is in turn defined by the
ability of such a system to maximize any possible reward metric that
one might care to supply to it.

Within the goals of AI research, this definition of intelligence has
been very productive.  It offers a clear metric of success that can be
measured in benchmarks, a metric which researchers can aim to improve.
The framing which equates more useful intelligence with better reward
maximization also plays an important role in broader work which
examines the alignment problem. Examples are the work of Omohundro
\cite{omohundro2008basic}, which uses the tools of economics and game
theory, and Bostrom's {\it Superintelligence}
\cite{bostrom2014superintelligence}, which adds the tools of
philosophy.  While this cross-disciplinary framing of intelligence as
reward maximization is useful, we feel it is also having an overly
narrowing effect on the debate.

Below, we will use the concept of cognitive architectures to reason
about the internals of intelligent entities, specifically internals
that combine reward maximization with further building blocks designed
to improve alignment.

Several parts of this paper somewhat fit a more general pattern in AI
scholarship.  The general pattern is that authors call for AI
researchers to extend, update, or even fully abandon their current
model of intelligence, in order to enable meaningful progress.
Historically, most of these calls have been motivated by the problem
of making progress in machine intelligence itself.  Increasingly,
there have been calls which are motivated by the problem of alignment.

We now review some of the diversity among these calls.  Russell
\cite{russell2019human} calls for a reshaping of the foundations of
the AI field away from reward maximization, and towards the idea that
the AI must be {\it beneficial}.  Amodei et al.\
\cite{amodei2016concrete} call for more research on a range of
open safety problems in machine learning, without going so far as to
call for a shift in foundations.  Dafoe et al.\
\cite{dafoe2021cooperative} call for scientists to reconceive
artificial intelligence as deeply social.  This means making a shift
towards research on cooperation in AI-AI and human-AI systems, and
away from AIs that operate in isolation, or aim to win zero-sum games
like chess.  Like most researchers currently working on AI alignment,
we feel that all these diverse routes have promise.  One aim of this
paper is to develop additional routes.

This paper does not fit the above pattern of calls in one important
way.  We are not directing our call towards AI researchers, but to all
participants in the current multi-disciplinary and multi-stakeholder
AI alignment debate.  Our aim is to improve the AI debate.  We see
improved debate as the main device for improving current and future
deployed AI technology.


\subsection{Cognitive architectures}
\label{introcogarch}

Kotseruba and Tsotsos \cite{Kotseruba201840YO} discuss how the term
cognitive architecture has historically been used to describe the
structure of both human and machine minds.  The term is often
associated with work in neuroscience that seeks to reverse-engineer
the human mind, but we will not proceed along these lines in this
paper.  For our purposes, we define a version of the term which we
will apply to humans, companies, governments, and autonomous AIs
alike.

First, we define a {\it cognitive process} as the full process inside
any such entity that perceives the world around it, further processes
these perceptions, and then initiates actions based on this
processing.  We then define a {\it cognitive architecture} as the set
of interconnected building blocks that make up the internals of a
cognitive process.

In the case of human organizations, we can describe their cognitive
architecture as one that was set up to initiate appropriate actions by
having human cognition interact with a set of written and unwritten
rules and goals.  Organizations increasingly use autonomous AI systems
to take some actions automatically, on a massive scale.  In our
terminology, these AIs become part of the organization's cognitive
architecture, but they also have cognitive architectures themselves.

If the business model of a company, or the entire market the company
operates in, is somewhat unaligned with human goals and needs, then we
can hardly expect that any AI deployed by the company will be better
aligned.  So demands made in an alignment debate are sometimes best
expressed as demands on business models or market regulators, not as
demands on a specific AI.

As a framing device, cognitive architectures have several uses in the
debate.  First, when a discussion gets stuck on the problems of reward
maximization, a reframing in terms of cognitive architectures might
help to overcome this barrier.  A second major use is to improve the
debating power of non-specialist stakeholders compared to that of
specialists, and those who might seek to creatively quote or misquote
specialists.

\subsection{The role of multi-stakeholder debate in alignment}

Multi-stakeholder debate is a tool that can be used to efficiently
collect relevant information, to compare proposed solutions, and to
trigger creative thinking.  But we also value this type of debate for
another reason.  We value its potential use as a social coordination
ritual.  This ritual can define, and create legitimacy for, agreed-on
actions and meanings.

Like Gabriel \cite{gabriel2020artificial}, we observe that the word
{\it aligned} encodes a moral judgment, and that it is therefore
impossible to define a single objectively and universally true meaning
for the word. For the purpose of this paper, we treat the meaning of
the word {\it aligned} as being preferably defined by a process of
informed debate and consensus building between representatives of all
affected human stakeholders.  The successful creation of a consensus
will lend a certain degree of moral legitimacy to the agreed-on
meaning of the word.  The consensus does this by defining a social
contract that specifies rights and obligations for all parties. In
particular, the contract will encode the moral right to claim that, if
certain obligations are met, the AI being created is an aligned AI.

Multi-stakeholder debate and social contract creation can also work to
solve collective action problems.  One example is the suppression of
AI arms race dynamics, that would otherwise create an unwanted
incentive for the participants in the race to compromise on AI safety
and quality control.

Of course, it is to be expected that any social contract will be
somewhat inexact, and that it may not always correctly
anticipate and cover all possible future developments.  When a social
contract is not updated in light of new developments, stakeholders may
soon stop accepting the implied definition of moral legitimacy.  The
same problem occurs when the contact is unenforceable or unenforced.
To be successful, participants in the alignment debate will need to
resolve many complex issues.  This paper examines just three of them
in detail.

In the framing of AI research, the ability to negotiate better social
or commercial contracts would be a clear sign of higher intelligence.
However, in this paper we are not considering the research goal of
making AIs better at negotiating contracts among themselves or with
humans.  We focus instead on contributing to the problem of making
humans better at negotiating certain broad social contracts among
themselves, better while using a process where the AIs do not
have any seat at the table.

If AIs which approximate or exceed full human intelligence are ever
developed, then both moral and practical arguments could be made for,
but also against, giving them a seat at the table above.  Bostrom
\cite{bostrom2014superintelligence} examines
this potential long term scenario in more detail.  Here, we do not
examine it any further.

\subsection{Structure of this paper}

This paper proceeds as follows.  Section 2 has general background
information on fashions in AI research.  Section 3 reviews the
cognitive architecture of a pure reward maximizer.  Section 4 then
modifies this architecture, adding elements not related to reward
maximization, to construct a law-abiding reward maximizer.  The
section also discusses how participants in debate can compensate for
the narrative pull of modern fashions in machine learning, if they
intend to examine or demand the creation of lawful AIs.

Section 5 examines reinforcement learners, a currently very popular
framework in both technical and policy discussions about autonomous AI
systems.  The section develops a detailed non-mathematical picture of
the cognitive architecture used by reinforcement learners. This
picture departs from how reinforcement learners are usually described
in the debate, by emphasizing that all reinforcement learners will
automatically construct and use some type of predictive world model.

Section 6 then uses this picture to explore some specific risks of
deploying highly advanced AIs in the current market system.  It
considers how these risks could be managed by creating a consensus
leading to regulatory action, where the regulator imposes certain
technical requirements on powerful market-facing AIs.  One promising
but currently unfashionable technical option, which is also explored
further in section 8, is that the regulator requires that the AI uses
a specifically incorrect world model.

Section 7 explores the field of machine learning in more detail, to
determine how participants in the debate should calibrate their
interactions with machine learning researchers.  Section 8 also
explores this topic, for the case of demanding specifically incorrect
world models.  Section 9 concludes.

\section{Fashions in AI}

We now make some remarks on terminology, acronyms, and fashion.
We use the term {\it machine learning} (ML) to denote a sub-field of
{\it artificial intelligence} (AI), where AI itself is a branch of
{\it information technology} (IT).  A notable linguistic development
is that all of fundamental and applied AI research is by now commonly
referred to as being ML research.  This creates a certain narrative
pull. When investigating an open problem related to AI, the first
impulse will be to look for a solution that leverages machine learning
techniques.

Among AI researchers, the term {\it reinforcement learning} (RL)
usually refers to the specific type of machine learning discussed by
Sutton and Barto \cite{suttonbarto}.  But the term has also acquired a
much broader meaning, in a linguistic drift process that often happens
when technologies trigger Gartner's commercial hype cycle
\cite{gartner}.  The acronym RL is increasingly being used to
denote any AI system designed to take autonomous actions towards a
goal, in a real or simulated physical environment.

When we talk about fashion in this paper, we will mostly describe the
prevailing fashions in ML and RL research.  But what is fashionable in
fundamental RL research, as described by Sutton \cite{sutton_bitter},
may not be appealing at all to an applied robotics researcher like
Brooks
\cite{brooks_better}.  It is the fashions and preferences among ML
and RL researchers which are having the greater impact on the current
debate.

\subsection{Good Old Fashioned AI}

Among AI practitioners, there is a useful acronym that denotes the
opposite of current hypes and fashions: GOFAI or Good Old Fashioned
AI.  A system that uses GOFAI will incorporate tried-and-tested AI
techniques which are no longer at the forefront of AI research.  One
reason to use GOFAI may be that the old technique simply has superior
performance for the job at hand.  Another may be that the old
technique makes safety engineering and worst-case risk analysis much
more tractable.

We feel that GOFAI is too often overlooked in the current debate about
AI safety and alignment.  There is a tendency to focus too much on the
features, limitations, and unknowns associated with the latest and
most fashionable ML techniques.

It is also necessary to apply a GOFAI lens to the many
recent AI strategy announcements by companies and governments,
announcements which often mention huge sums of money going to more AI.
If GOFAI is overlooked by a participant in the debate, they may easily
get the impression that these players have committed to an
irresponsible rush to apply the latest hyped AI technologies to every
aspect of human life.

GOFAI is definitely not being overlooked in the near-term alignment
debate about AI-based fairness and discrimination. If a regulator
requires that a company must be able to explain to a job applicant
just how it is avoiding any unfair bias in its decision support
systems, then this often forces the use of GOFAI instead of modern
deep neural nets.  Unlike current deep neural nets, GOFAI techniques
can produce systems with internal moving parts that can be tractably
audited and explained.

\section{Pure reward maximizers}

A {\sl pure reward maximizer} is a cognitive architecture that is
fully devoted to taking actions which maximize an expected future
reward.  In AI research and in game theory, this reward can be any
metric of success or merit we may care to define.  In economic
theory, when applied to the cognitive architecture of companies, the
reward metric is most often company profits.

In the medium and long-term AI alignment debate, it is often useful to
draw parallels between a pure reward maximizing AI and a company that
cares for nothing but profits. This can uncover many AI safety issues
and failure modes that need to be addressed.  But our goal here is to
consider how, after having such a discussion about pure reward
maximizers, one might move forward.

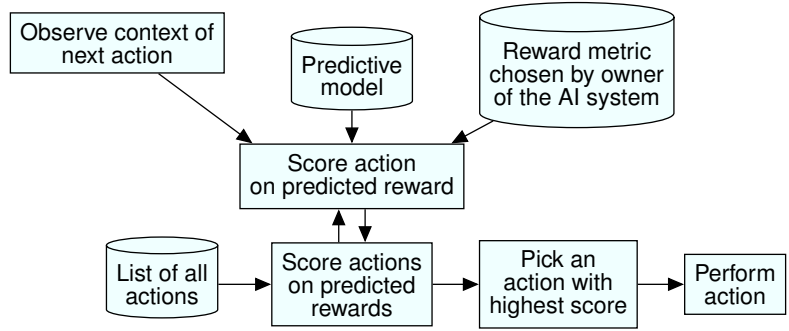
\begin{figure}[h]
\centering
\resizebox{0.75\textwidth}{!}{
    \begin{tikzpicture}[
    node distance = 5mm, line width=0.20mm,
 disc/.style = {shape=cylinder, draw, shape aspect=0.15,
                shape border rotate=90, inner sep=1.5mm,
		align=center, font=\linespread{0.9}\selectfont,fill=lightblue},
  alg/.style = {rectangle, draw, align=center, font=\linespread{0.9}\selectfont,fill=lightblue,inner sep=1.5mm},
  xtra/.style = {fill=lightgreen},
                    ]
\sf
\node (n5) [disc]  {Predictive\\ model};
\node (score1a) [alg, below=5mm of n5]  {Score action\\ on predicted reward};
\edge{n5}{score1a};
\node (scorea) [alg,below=5mm of score1a]  {Score actions\\ on predicted\\ rewards};
\edge[transform canvas={xshift=-2mm}]{scorea}{score1a};
\edge[transform canvas={xshift=2mm}]{score1a}{scorea};

\node (picka) [alg, right=7mm of scorea]  {Pick an\\ action with\\ highest score};

\node (performa) [alg, right=7mm of picka]  {Perform\\ action};

\node (loa1) [disc, left=8mm of scorea] {List of all\\ actions};

\edge{loa1}{scorea};
\edge{scorea}{picka};
\edge{picka}{performa};

\node (rewfunc) [disc, right=10mm of n5,yshift=0mm] {Reward metric\\ chosen by owner\\ of the AI system};
\path (rewfunc) edge[->] ([xshift=-2mm]score1a.north east);

\node (obs) [alg, left=10mm of n5,yshift=5mm] {Observe context of\\ next action};

\path (obs) edge[->] ([xshift=2mm]score1a.north west);

\end{tikzpicture}
}
\caption{
Graphical depiction of the cognitive architecture of a pure reward
maximizing AI, using a data-flow diagram.  Arrows represent data flow.
Each cylinder is an amount of data.
Rectangular boxes are pieces of software.}
\label{figrm}
\end{figure}

Figure \ref{figrm} depicts the cognitive architecture of a reward
maximizer as a set of interconnected building blocks.  The maximizer
first observes its environment to determine the context of the next
action to be taken. It then uses this context together with a
predictive model, to score a list of all possible actions which might
be taken on a reward metric.  After scoring each possible action on
the predicted reward, it picks and performs one of the actions which
have gotten the highest reward score.

In naming the individual building blocks, we have avoided the use of
specialist terminology as much as possible.  The intent is to make
this picture maximally useful as tool for clarifying and structuring
multi-disciplinary debate.

\section{Law-abiding reward maximizers}

Aligned commercial companies not only care about profits, they also
care about the law.  Inside the cognitive architecture of a
law-abiding company, we may find legal compliance officers.  Their
role is to examine if proposed company actions or policies would be in
violation of the law.  If so, they are supposed to block the use of
these proposed actions or policies, regardless of how this will affect
profits.

We could demand that the same cognitive architecture is also used in
an AI that makes autonomous decisions without human review.  To build
such an AI, we need a piece of software that automates the job of the
human compliance officer.  In basic IT terminology, this piece
software takes two inputs. First, it needs a description of the proposed
action. Second, it needs relevant contextual information, a
description of the target environment to which the proposal will be
applied.  The output would be a yes or no answer about whether the
proposed action is legal in that context.

There are several options for implementing the above compliance
officer software.  To further illustrate our point about GOFAI, we
consider the option of using an {\it expert system}.  Expert systems
are a type of AI technology that experienced peak hype in the 1980s.
An expert system takes some inputs and then applies certain
computer-readable rules to draw a conclusion about these inputs.  To
build the compliance officer expert system needed above, its
programmers would consult with the company's legal department to
locate all applicable laws that the AI might violate with its actions.
They then convert these laws into computer-readable rules for the
expert system.


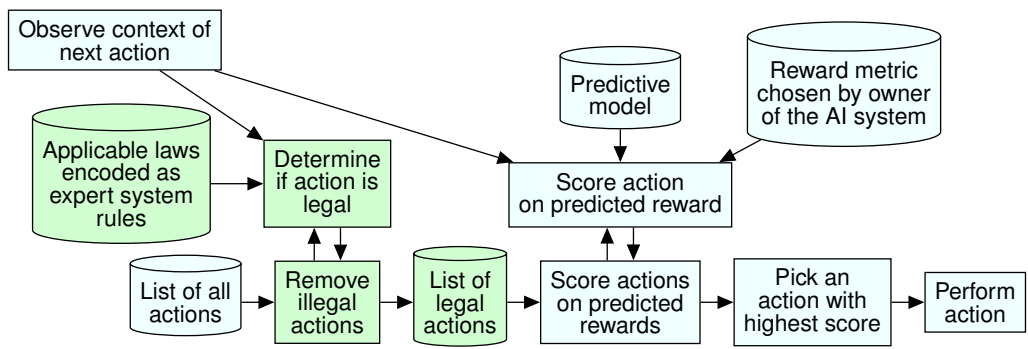
\begin{figure}[h]
\centering
\resizebox{0.98\textwidth}{!}{
    \begin{tikzpicture}[
    node distance = 5mm, line width=0.20mm,
 disc/.style = {shape=cylinder, draw, shape aspect=0.15,
                shape border rotate=90, inner sep=1.5mm,
		align=center, font=\linespread{0.9}\selectfont,fill=lightblue},
  alg/.style = {rectangle, draw, align=center, font=\linespread{0.9}\selectfont,fill=lightblue,inner sep=1.5mm},
  xtra/.style = {fill=lightgreen},
                    ]
\sf
\node (n5) [disc]  {Predictive\\ model};
\node (score1a) [alg, below=5mm of n5]  {Score action\\ on predicted reward};
\edge{n5}{score1a};
\node (scorea) [alg,below=5mm of score1a]  {Score actions\\ on predicted\\ rewards};
\edge[transform canvas={xshift=-2mm}]{scorea}{score1a};
\edge[transform canvas={xshift=2mm}]{score1a}{scorea};

\node (picka) [alg, right=5mm of scorea]  {Pick an\\ action with\\ highest score};

\node (performa) [alg, right=5mm of picka]  {Perform\\ action};

\node (loa2) [disc, xtra,left=5mm of scorea] {List of\\ legal\\ actions};

\node (filterl) [alg, xtra, left=5mm of loa2] {Remove \\ illegal\\ actions};

\node (loa1) [disc, left=5mm of filterl] {List of all\\ actions};

\node (filtera) [alg,xtra, above=5mm of filterl] {Determine\\ if action is\\ legal};

\node (xprules) [disc, xtra, left=8mm of filtera] {Applicable laws\\ encoded as\\ expert system\\ rules};

\edge[transform canvas={xshift=-2mm}]{filterl}{filtera};
\edge[transform canvas={xshift=2mm}]{filtera}{filterl};

\edge{loa1}{filterl};
\edge{filterl}{loa2};
\edge{loa2}{scorea};
\edge{scorea}{picka};
\edge{picka}{performa};
\edge{xprules}{filtera};

\node (rewfunc) [disc, right=10mm of n5,yshift=0mm] {Reward metric\\ chosen by owner\\ of the AI system};
\path (rewfunc) edge[->] ([xshift=-2mm]score1a.north east);

\node (obs) [alg, above=5mm of xprules,xshift=-1mm] {Observe context of\\ next action};

\edge{obs}{score1a};
\edge{obs}{filtera};

\end{tikzpicture}
}
\caption{
Cognitive architecture of a law-abiding reward maximizer.
To improve alignment, the pure reward maximizer from
figure \ref{figrm} is extended with additional green building blocks.}
\label{figlarm}
\end{figure}

The full design of the cognitive architecture of the demanded
law-abiding profit-maximizing AI might then look as follows (figure
\ref{figlarm}).  Whenever the AI has to decide on taking the next
autonomous action, it first uses a software module to construct a long
list of actions that could be taken.  For example, each action on that
list could be to show one specific advertisement, out of all the
available ads, to an end user.  The list of actions is then sent to
the expert system, which will remove all illegal actions from the
list.  It may for example be illegal to show ads about gambling to
minors.  The remaining actions on the list are scored by a different
subsystem, which estimates their profitability.  The AI will then pick
a legal action that has a maximal profitability score, and
autonomously perform this action.

Cognitive architectures like the above are routinely deployed.  ML
experts will usually describe them using a short mathematical formula
which fully captures the overall intent of the above information
processing chain.  If not pressed by other participants in the debate,
they may never switch to the more accessible descriptive language of
step-wise bureaucratic decision making we have used above.

We now examine a further issue.  If one were to pose the problem of
building a law-abiding reward maximizer to a modern ML researcher, it
is unlikely that they will immediately bring up expert system
technology.  Expert systems are not even ML systems.  They do not
learn because all their knowledge has to be carefully constructed by
hand, in the form of computer-readable rules.  This makes expert
systems unfashionable among ML researchers, and it also makes them
unfashionable in the applied AI community.  Among applied AI
programmers, few tasks would be considered less glamorous than the
task of hand-translating a body of law identified by the legal
department into expert system rules.

What is more likely to happen in an alignment debate is that the ML
researcher or AI practitioner will immediately express great
enthusiasm for addressing the problem of law-abiding AI with modern
machine learning technology.  This enthusiasm might take the form of a
proposal for a research project that examines how deep neural network
based natural language processing (NLP) can be used to automatically
convert a written body of law into computer-readable rules. The
proposed research project would investigate the unsolved problem of
making this fully automated conversion process work robustly in the
general case.

So it is easy for a discussion about AI and the law to follow a
narrative flow which arrives at the conclusion that certain reasonable
demands cannot be met by modern ML technology, unless further ML
research is successful in solving open problems.  This can put the
debate about meeting stakeholder demands into an undesirable holding
pattern.  If stakeholders want to overcome this obstacle to progress,
they can instead examine the cognitive architecture of a law-abiding
company, and then demand that GOFAI is used to replicate the same
architecture in the AI.

\section{Reinforcement learners}

In ML research, reinforcement learning \cite{suttonbarto} is
currently the most popular framework for considering autonomous AI
systems that interact with an environment over multiple time steps. We
now examine this framework in more detail.

A reinforcement learner is a cognitive architecture designed to
autonomously take actions over time.  The mathematical convention is
that a reinforcement learner takes one action per time step.  All of
these actions are supposed to contribute to a single goal, a goal
defined by a {\it reward function}.  If we use basic IT terminology to
describe it, then the reward function is a piece of software that will
be run inside the reinforcement learner at the end of each time step.
The reward function software will compute and deliver a single number,
which is interpreted by the cognitive architecture as a reward
received in that particular time step.  For example, the reward
function might query a database to determine the profit made by a
company during that time step, and deliver that profit as the reward.

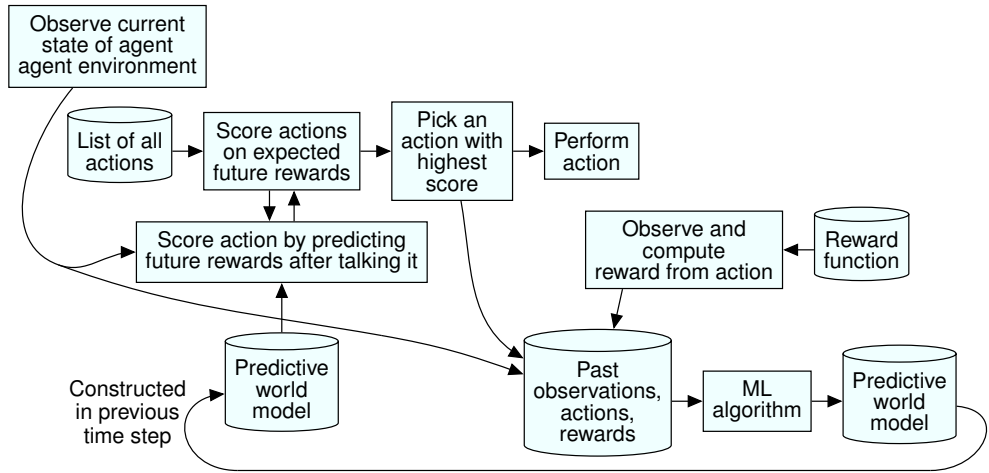
\begin{figure}[h]
\centering
\resizebox{0.98\textwidth}{!}{
    \begin{tikzpicture}[
    node distance = 5mm, line width=0.2mm,
  disc/.style = {shape=cylinder, draw, shape aspect=0.15,
                shape border rotate=90, inner sep=1.5mm,
		align=center, font=\linespread{0.9}\selectfont,fill=lightblue},
  alg/.style = {rectangle, draw, align=center, font=\linespread{0.9}\selectfont,fill=lightblue,inner sep=1.5mm},
  xtra/.style = {fill=lightgreen}
                    ]
\sf
\node (n5) [disc]  {Predictive\\ world\\ model};

\node (n1) [disc, right=30mm of n5,yshift=-2mm] {Past\\ observations, \\ actions,\\ rewards};
\node (n2) [alg, right=5mm of n1]  {ML\\ algorithm};
\node (n3) [disc,right=5mm of n2]  {Predictive\\ world\\ model};

\edge{n1}{n2};
\edge{n2}{n3};

\node (score1a) [alg, above=8mm of n5]  {Score action by predicting\\ future rewards after talking it};
\edge{n5}{score1a};
\node (scorea) [alg, above=5mm of score1a]  {Score actions\\ on expected\\ future rewards};
\edge[transform canvas={xshift=-2mm}]{scorea}{score1a};
\edge[transform canvas={xshift=2mm}]{score1a}{scorea};

\node (picka) [alg, right=5mm of scorea]  {Pick an\\ action with\\ highest\\ score};

\node (performa) [alg, right=5mm of picka]  {Perform\\ action};

\node (loa1) [disc, left=5mm of scorea] {List of all\\ actions};

\edge{loa1}{scorea};
\edge{scorea}{picka};
\edge{picka}{performa};

\node (obs) [alg, above=3mm of loa1,xshift=-2mm] {Observe current\\ state of agent \\ agent environment};

\node (comprew) [alg, below=5mm of performa, xshift=15mm] {Observe and\\ compute\\ reward from action};
\node (rewf) [disc,right=5mm of comprew] {Reward\\ function};

\path ([xshift=-10mm]comprew.south) edge[->] (n1);
\edge{rewf}{comprew};

\path (picka) edge[->,out=280,in=150] (n1);

\node (hp1) [draw=none, below=12mm of loa1, xshift=-10mm] {};
\node (hp2) [draw=none, right=50mm of hp1, yshift=-11.5mm] {};

\path (obs) edge[-,out=230,in=160] (hp1.center);
\path (hp1.center) edge[->,out=-20,in=180] (score1a);
\path (hp1.center) edge[-,out=-20,in=170] (hp2.center);
\path (hp2.center) edge[->,out=-10,in=160] (n1);

\node (hp3) [draw=none, below=3mm of n3, xshift=5mm] {};
\node (hp4) [draw=none, left=110mm of hp3] {};

\path (n3) edge[-,out=-5,in=0,looseness=2] (hp3.center);
\path (hp3.center) edge[-,out=180,in=0,] (hp4.center);
\path (hp4.center) edge[->,out=180,in=185,looseness=2] (n5);

\node (txt2) [rectangle, draw=none, align=center, left=5mm of n5,yshift=-4mm] {Constructed\\ in previous\\ time step};

\end{tikzpicture}
}
\caption{
Cognitive architecture of a generic reinforcement learner.  Many
popular specific reinforcement learning architectures merge or approximate the
building blocks depicted, to speed up computations or to save on
memory usage.  To simplify the presentation, we have not included any
building blocks that trigger `exploration' actions, even though most
reinforcement learning systems have them.}
\label{figrl}
\end{figure}

As shown in figure \ref{figrl}, the cognitive architecture of a
reinforcement learner incorporates machine learning in order to
predict the general relation between the possible actions it could
take and its future rewards.  It will use these predictions to always
take an action that will maximize a weighted sum of all expected
future rewards.  Reinforcement learners may operate for many time
steps before receiving the first non-zero reward.

Like humans, governments, and businesses, reinforcement learners have
a cognitive architecture capable of long-term planning, of making
investments now to capture larger rewards in future.  They can be
surprisingly creative in finding ways to maximize their summed reward.
They may find strategies which surprise and occasionally disappoint
those who designed the reward function, as memorably discussed by
Clark and Amodei \cite{boat}. 
Reinforcement learners are not the only type of AI capable of
potentially dangerous creativity, for example Lehman et al.\
\cite{lehman2020surprising} show that the same problem also exists
with evolutionary machine learning.

\subsection{Automatically constructed world models}

A reinforcement learner is designed to automatically and autonomously
construct a {\it world model} it will use for decision making, to
construct it via machine learning.  In general IT terminology, the
world model constructed by the reinforcement learner is a digital
model that allows it to make predictions about the future, based on
information about the past and present.

Usually, the learned digital world model will allow the reinforcement
learner to predict, for each alternative action it might take next,
how taking this action will impact its expected summed future rewards.
But in some designs, for example when using policy gradient methods,
the predictive model being constructed can only be used to compute an
estimate of the best next action, it cannot score each individual
action.
The term {\it model-free} reinforcement learner is commonly used for
reinforcement learning architectures which do not aim to construct the
type of rich predictive world model which can project entire future world
states.  Policy gradient learners are examples of model-free
reinforcement learners.  The common framing is that a model-free
reinforcement learner constructs a limited predictive function only,
not a fully realized predictive world model.  But for our purposes, we
will interpret even these limited functions as being predictive world
models produced by machine learning.  They still encode predictions of
how the world will mediate between action and summed future reward.

The automatic building of world models by reinforcement learners has
obvious economic advantages.  It can replace the GOFAI technique of
having a team of domain experts and programmers carefully build a
world model by hand.  But beyond mere economics, there is a stronger
force that drives the enthusiasm for reinforcement learners.
Reinforcement learning research is driven in part by a vision that is
common among IT professionals, by the idea it would be great to
automate away repetitive tasks like building new domain specific world
models by hand.  The pursuit of this vision by ML researchers has
created some impressive results.  For an increasing number of
applications, modern machine learning can automatically build world
models that will allow the AI to massively outperform earlier AIs
using carefully hand-crafted world models.  A lot of modern ML
research is focused on pushing the edge forward even further.
 
\section{Risks from cheap and accurate world models}

We now use a thought experiment to explore some of the consequences of
using ever more accurate world models in automated decision making.

Say that a company, operating in some market, wants to fully automate
certain customer interactions.  They set up an advanced reinforcement
learner that their customers will interact with via a web site.  They
configure the reinforcement learner to maximize company profits.  Now,
if customers feel mistreated by the actions of the reinforcement
learner, they might contact the market regulator, who may impose a
fine that will lower company profits.

Consider a reinforcement learner which automatically builds a world
model that will start to correctly represent the above mechanisms by
which profit is determined.  The system will learn to estimate the
exact costs of treating certain customers in a certain way. It will
learn to deliver correct but particularly expensive services only to
those customers who are likely to complain to the regulator when these
services are withheld.  To maximize overall profits. customers who are
less likely to cause fines when mistreated will be treated less
correctly.  The system has a cognitive architecture which is
configured to treat an expected fine as a price, as a cost of doing
business, not as a social signal that it must absolutely try to avoid
treating any customer that way in future.

This leads to a moral question, the question whether the above form of
gaming the regulatory system is aligned.  Different stakeholders may
answer this question differently.  A first stakeholder may feel that
treating a fine as a price, or discriminating between customers in
this way, is always morally wrong.  A second stakeholder may feel that
this emergent AI behavior can still be morally acceptable, as long as
even the worst cases of customer treatment delivered still fall
within a range of accepted business practices.  Stakeholder debate
will be needed to resolve this moral question for society.

We now proceed with a further step in the thought experiment.
Consider what might happen in a future where every company and
customer in every market has the ability to cheaply build and deploy
these gaming capable AIs.  If no government steps in to intervene, we
can expect that in every market, commercial pressures will create a
race to the bottom in AI-controlled gaming.  To stay in business,
every company will ever more desperately use its automated systems to
game and counter-game everybody and everything around it.  This gaming
could easily destabilize any market, and society as a whole.

Zubov has argued \cite{zuboff2019age} that such a computer-aided
breakdown of the traditional market system has already happened.  We
believe instead that the highly effective automated gaming of customers and
regulatory structures, and the race to the bottom scenario above,
would require at least one more major technical breakthrough.

As the timing of technical breakthroughs is unpredictable, there are
good reasons for already having a debate about the possibilities for
government regulation, to prevent the above race to the bottom.  We
believe that most companies would greatly prefer to operate in markets
that reward honest interactions with customers and other stakeholders
more than they reward gaming.  So on the political side of the debate,
the problem of building a consensus for the enforcement of certain
regulatory demands that constrain automated gaming by AIs should be
tractable.  This leaves the question of what these demands should look
like technically.

\subsection{Technical options for the regulator}

The law-abiding cognitive architecture discussed earlier may go some
way towards suppressing gaming, depending on how good the applicable
laws are.  We now list two further ways to address the problem of
gaming by autonomous AIs, to address it in a way that could be audited
by a market regulator.

\begin{figure}[t]
\centering
\resizebox{1\textwidth}{!}{
    \begin{tikzpicture}[
    node distance = 5mm, line width=0.2mm,
  disc/.style = {shape=cylinder, draw, shape aspect=0.15,
                shape border rotate=90, inner sep=1.5mm,
		align=center, font=\linespread{0.9}\selectfont,fill=lightblue},
  alg/.style = {rectangle, draw, align=center, font=\linespread{0.9}\selectfont,fill=lightblue,inner sep=1.5mm},
  xtra/.style = {fill=lightgreen}
                    ]
\sf
\node (n5) [disc,xtra]  {Specifically incorrect \\ predictive\\ world model};

\node (n1) [disc, right=30mm of n5,yshift=-2mm] {Past\\ observations, \\ actions,\\ rewards};
\node (n2) [alg, right=5mm of n1]  {ML\\ algorithm};
\node (n3) [disc,right=5mm of n2]  {Correct\\ predictive\\ world\\ model};
\node (n4) [alg,xtra,right=5mm of n3] {Model\\ Editing};

\edge{n1}{n2};
\edge{n2}{n3};
\edge{n3}{n4};

\node (n4b) [disc,xtra,above=5mm of n4] {Desired\\ incorrect\\ model\\ elements};

\edge{n4b}{n4};

\node (score1a) [alg, above=8mm of n5]  {Score action by predicting\\ future rewards after talking it};
\edge{n5}{score1a};
\node (scorea) [alg, above=5mm of score1a]  {Score actions\\ on expected\\ future rewards};
\edge[transform canvas={xshift=-2mm}]{scorea}{score1a};
\edge[transform canvas={xshift=2mm}]{score1a}{scorea};

\node (picka) [alg, right=5mm of scorea]  {Pick an\\ action with\\ highest\\ score};

\node (performa) [alg, right=5mm of picka]  {Perform\\ action};

\node (loa1) [disc, left=5mm of scorea] {List of all\\ actions};

\edge{loa1}{scorea};
\edge{scorea}{picka};
\edge{picka}{performa};

\node (obs) [alg, above=3mm of loa1,xshift=-2mm] {Observe current\\ state of agent \\ agent environment};

\node (comprew) [alg, below=5mm of performa, xshift=15mm] {Observe and\\ compute\\ reward from action};
\node (rewf) [disc,right=5mm of comprew] {Reward\\ function};

\path ([xshift=-10mm]comprew.south) edge[->] (n1);
\edge{rewf}{comprew};

\path (picka) edge[->,out=280,in=150] (n1);

\node (hp1) [draw=none, below=12mm of loa1, xshift=-10mm] {};
\node (hp2) [draw=none, right=50mm of hp1, yshift=-11.5mm] {};

\path (obs) edge[-,out=230,in=160] (hp1.center);
\path (hp1.center) edge[->,out=-20,in=180] (score1a);
\path (hp1.center) edge[-,out=-20,in=170] (hp2.center);
\path (hp2.center) edge[->,out=-10,in=160] (n1);

\node (hp3) [draw=none, below=4mm of n4, xshift=5mm] {};
\node (hp4) [draw=none, left=152mm of hp3] {};

\path (n4) edge[-,out=-5,in=0,looseness=2] (hp3.center);
\path (hp3.center) edge[-,out=180,in=0,] (hp4.center);
\path (hp4.center) edge[->,out=180,in=185,looseness=2] (n5);

\node (txt2) [rectangle, draw=none, align=center, left=5mm of n5,yshift=-4mm] {Constructed\\ in previous\\ time step};

\end{tikzpicture}
}
\caption{
Design of a more aligned reinforcement learner.  A regulator might
demand that the green building blocks are added to ensure that a
specifically incorrect world model is used, one that produces
automated decision making more in line with agreed-on human goals and
needs.}
\label{figrlwm}
\end{figure}
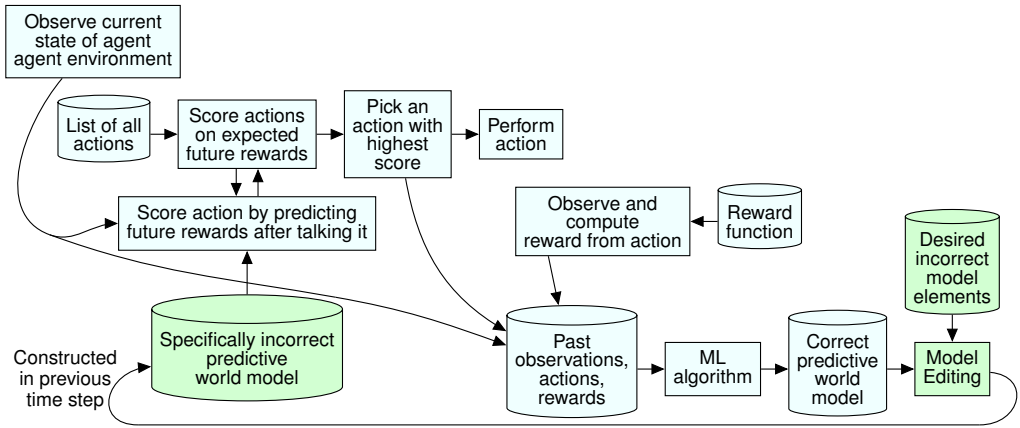

\begin{enumerate}
\item {\bf Limited or specifically incorrect world models.}
Going back to the example of an AI gaming the regulatory system, if we
can remove the regulator from the AI's world model, we will break the
connection between the existence of the regulator and the expected
profits that the AI's world model will project. This will make the AI
lose both the incentive and the ability to treat a fine as a price.

Figure \ref{figrlwm} shows a cognitive architecture which ensures the
use of a specifically incorrect world model to drive decision making.
Two general questions are raised when we consider the green building
blocks which have been added.  First, if we want to improve alignment,
what kind of world model imperfections might usefully be imagined and
demanded?  Second, what is the technical feasibility of the depicted
{\it model editing} operation?  We will examine both questions in more
detail in section
\ref{limited}.

Moving away from figure \ref{figrlwm}, we can also consider another
type of design intervention, one that combines the earlier generic RL
architecture from figure \ref{figrl} with a reward function that has
specific imperfections designed in.  In the gaming example, the
company might combine a generic RL architecture with a reward function
that excludes the impact of all fines from the profit calculation.  We
can interpret the resulting RL system as a profit maximizer which has
an incorrect model of how profit can be maximized.  This incorrect
model will greatly suppress the disparate treatment of customers we
described.

Note that this approach might not perfectly erase the regulator from
the learned model.  It is likely that the presence of the regulator
has secondary effects beyond mere fines, and these secondary effects
might still be correctly captured by the model.  But we are not
necessarily looking for the mathematically perfect removal of all
gaming incentives from all AIs.  Markets and society are robust enough
that they can tolerate a certain background level of gaming.  Certain
types and levels of gaming may even be explicitly agreed on to be
morally admissible, because they add more color to life.
\item {\bf More aligned reward functions.}
We could also consider building a reward function which rewards not
only an economic goal, but also general pro-social behavior.  The
reward function may combine a profit measure with other metrics that
detect and penalize socially unaligned behaviors like gaming, breaking
laws, manipulating human oversight, or deceit in general.

In the context the above example, we could add a term to the reward
function that multiplies any fine received by a factor of 100.  This
will produce a cognitive architecture that is much more reluctant to
trigger fines, which would make the AI more aligned, but arguably only
if all mistreated customers are equally likely to contact the
regulator.  A safer option would be for the system designers to
hand-code a penalty term that more directly measures an agreed-on
metric of incorrect treatment.
\end{enumerate}

Writing pro-social reward functions for powerful AIs will not be easy.
In reviews of AI safety problems written by ML researchers, for
example in Amodei et al.\
\cite{amodei2016concrete}, it is common to find the observation that a
powerful AI cannot not be fully safe or aligned if its reward function
is incorrect, if it leaves loopholes in the encoding of desired
behavior.  The AI alignment literature often proceeds to identify the
problem that fallible humans cannot ever be expected to write a fully
correct reward function, definitely not if this reward function has to
encode the full goals and needs of humanity.  However, ML and
alignment researchers seldom proceed from these observations to work
on the problem of writing down a reward function which is at least as
correct as humanly possible.  In the next section, we examine why this
might be the case.

Overall, we feel that the broad topic of pro-social reward function
design needs more attention, and that it needs to become more
fashionable among AI alignment technology researchers.  Some existing
work can be found in Turner et al. \cite{turner2020conservative},
Krakovna et al.\ \cite{krakovna2020avoiding}, Soares et
al. \cite{corr}, Holtman \cite{holtmanitr}, and Vamplew et al.\
\cite{vamplew2018human}.
Of course, an act of reward function design necessarily has to happen
whenever a business wants to deploy a modern autonomous AI.  When
challenged, many businesses would claim that they have indeed written
and deployed pro-social reward functions.  But as long as these
businesses treat their reward functions as trade secrets, this is of
no great help to open scholarship or open debate.

\section{Preferences and deflection among ML researchers}

Russell \cite{russell2019human} describes how Bostrom's call
\cite{bostrom2014superintelligence} for deep
thinking about long-term alignment led to a `not so great debate' in
the AI community.  Notably, well-known AI researchers have made
`instantly regrettable remarks' to deflect such calls.  Russell
identifies some of the drivers that may sustain the not-so-great
nature of the debate, like the possible fear of losing research
funding, and false dichotomies hardening into tribalism.  Baum
\cite{baum2018reconciliation} and Stix and Maas
\cite{stix2021bridging} show that tribalism is being sustained
by the difference in near-term versus long-term focus.  They consider
how to bridge this divide, so that the debate can move more smoothly
forward.  Here, we consider a different driver which creates a divide
that is less easily bridged.

There is a division of labor that exists in all of information
technology.  It is common in IT to split the problem of making a
computer do something new and useful over two teams: the {\it
specification team} and the {\it implementation team}.  The job of the
specification team is to write an unambiguous specification of the
useful thing that the computer should do. This specification is then
handed to the implementation team, which will figure out how to make a
computer do this useful thing in the most efficient way.

In this division of labor, only the specification team will interact
with the stakeholders affected by the implementation. It is the job of
the specification team to correctly identify and triangulate
stakeholder needs.  When the stakeholders concerned have largely
conflicting goals or tense relations, the specification team will have
to navigate and resolve many `people problems', if they are to produce
something truly useful.  The specification being developed will
inevitably be judged by all stakeholders on how the resulting computer
system will affect the balance of power and social contracts between
them.  The specification team may sometimes succeed in defusing tense
stakeholder relations by locating a proposal that will be perceived as
a win-win improvement by all.  But such a happy outcome is by no means
guaranteed.

Many IT technologists would prefer to have a career path which lets
them forever avoid working on the people problems and
stakeholder tensions that the specification team has to handle.  They
prefer a path that puts them firmly in the implementation team only,
or in a position where their only concern is to deliver useful tools
and technologies to the implementation team.  This desire is by no
means universal, some technologists may explicitly seek out a career
in the specification team, others may seek the variety that comes from
being on both teams.

We now turn to the career path offered by ML research.  Current
mainstream ML research treats the AI's reward function as the sole
specification of the useful thing an autonomous AI should do.  This
means that the specification team problem of writing a useful or
aligned reward function is out of scope for the ML researcher.  The ML
researcher is concerned instead with the mathematically clear and
unambiguous problem of building technology that will learn to maximize
any reward function one might supply.  Progress on this problem will
be judged by objective machine learning benchmarks.  All these things
combine into the promise of a politics-free career path with a clear
and level playing field.

Those working on AI alignment have by now written many papers that aim
to introduce ML researchers to the methods and tools used by the
specification team.  Some notable examples are the discussion of
incomplete contracting by Hadfield-Menell and Hadfield
\cite{hadfield2019incomplete}, and of 
incompletely theorized agreements by Stix and Maas
\cite{stix2021bridging}.
Gabriel \cite{gabriel2020artificial} offers an accessible discussion
of how fair principles for AI alignment could be identified.  Dobbe et
al.\ \cite{dobbe2021hard} present a specific process framework that
could be used by AI designers to make hard choices about the AI's
sociotechnical impact, based on developing and shaping stakeholder
feedback channels.  Selbst et al.\ \cite{selbst2019fairness} consider
the case of fairness in ML.  More generally, the fields of software
engineering, systems engineering, and science and technology studies
have developed and documented many useful specification team tools.

Our personal experience is that, when one engages with an individual
technical expert and gently tries to push them into picking up these
tools to join the specification team, the vast majority of these
experts will respond by ignoring, countering, or deflecting the
applied pressure. \label{pressure}

We expect that any alignment project or funding effort which aims to
make more ML researchers interested in contributing to all parts of
the alignment problem will inevitably run into this barrier.  One
might try to overcome it by finding more clever ways to push harder,
but we feel that this approach is both unkind and unlikely to produce
sufficient results.

The more productive option is to route around this barrier, to endorse
and promote a division of labor which does not expect ML researchers
to lead every charge.  This means that one should avoid a framing
where all of AI alignment research is, or must become, a sub-field of
AI research.  A better way forward is to declare that many of the
problems in AI alignment are broad political and systems engineering
problems, not ML problems.  Here, we use the term systems engineering
to denote a multidisciplinary field that contains the entire path from
specification to implementation inside of its scope.  Furthermore, the
system being engineered is not just the technical artifact itself, but
the entire set of interactions between the artifact and broader
society.

That being said, we also note that AI safety agendas like Amodei et
al.\ \cite{amodei2016concrete} list many important problems for which
progress can be made while staying within the traditional scope of ML
research and of the implementation team.  In the ML research
community, there is considerable enthusiasm for working on many of
these open safety problems.

\subsection{Machine learning of an aligned reward function}

There is also enthusiasm among ML researchers for the option of
creating a pro-social reward function automatically, via machine
learning.  This type of automation removes programmer labor, but does
require the presence and involvement of a human teacher.  The
convention is that the reward function to be learned already exists in
the teacher's head.  The teacher will communicate this function to the
AI by interacting with it.

To make a reinforcement learner automatically learn the teacher's
reward function, we could create a setup where the reward function
software inside its cognitive architecture determines the numerical
reward for each time step by reading out a remote control device held
by the teacher.  The teacher has two buttons: one to signal approval
of past behavior by giving a positive reward in the current time step,
one to punish past behavior with a negative reward.  If the teacher
rewards pro-social behaviors and punishes non-social behaviors, the
learned reward function will be a pro-social one.  Beyond
reinforcement learning, another approach to learning a reward function
from a human teacher is {\it inverse reinforcement learning} as
described by Ng and Russell
\cite{ng2000algorithms}.

For use cases where this works well, the above techniques have obvious
economic benefits.  They remove the need for a programmer to hand-code
a reward function after interviewing the teacher themselves.  But we
believe that this does not fully explain the enthusiasm for reward
function learning among many of the IT technologists participating in
the AI alignment debate.  This enthusiasm is better explained by the
hope that reward function learning can be scaled up to automate away
all the difficult stakeholder interactions that these technologists
imagine they must otherwise engage in themselves, if they want to
create a more stakeholder-aligned AI.

Long-term alignment discussions often consider the many dangers and
failure modes inherent in the above remote control based setup.  One
commonly mentioned failure mode is that the AI may creatively use
violence to force the teacher into forever pressing the approval
button.  Technical research on the long-term alignment problem is by
now mostly concerned with designing and studying very different setups
that suppress or remove such failure modes.
For technical readers, some examples of alternatives are in
Hadfield-Menell et al.\
\cite{hadfield2016cooperative}, Everitt et al.\
\cite{Everitt2019-3}, Orseau and Armstrong
\cite{orseau2016safely}, Holtman \cite{holtmanitr},
Cohen et al.\ \cite{cohen2020asymptotically},
Armstrong et al.\ \cite{armstrong2020pitfalls}, and Drexler
\cite{drexler2019reframing}.

\section{Limited or specifically incorrect world models}
\label{limited}

The cognitive architecture of the human mind has a great ability to
take actions based on limited or specifically incorrect world models,
Even small children can play a game where they all pretend to be
pirates while the floor is lava.  As the game goes on, they will often
negotiate further rules among themselves to keep things more balanced
and interesting.

In adult life, we may demand that a business owner will treat male and
female customers exactly alike, even when they know full well that one
of these customer types is more profitable.  We do not doubt that the
business owner has the mental ability to meet this demand if they want
to.  To frame this demand in terms of a cognitive architecture, we
expect that the business owner, while maximizing profit, will take
certain decisions by using a world model in which the distinction
between genders is erased.  The modern concept of fair and equal
treatment of citizens by government can also be expressed as a demand
that civil servants must make decisions without taking certain facts
into account.

When one examines various social contracts, one can note that many
demands made in them are demands that certain players must use
specifically limited world models when making decisions which affect
others.  For powerful players like governments, it is common to demand
especially severe limitations.  So we may demand that such limitations
are also present in the cognitive architecture of any powerful AI that
interacts with humankind.

Overall, we feel that a broad range of useful AI alignment demands can
be developed, clarified, and expressed as demands that the AI world
model incorporates some specifically designed imperfections.  We feel
that this is a promising but still largely overlooked direction for AI
research, and for advancing the alignment debate.

The sub-field of AI fairness research already concerns itself with the
specification and construction of desired imperfections in predictive
models, though it usually frames these as being improvements, not
imperfections.  Christian
\cite{christian2020alignment} has an accessible discussion of the
recent developments in this field, and shows how AI fairness research
has uncovered some surprising difficulties.  The seemingly simple idea
of a predictive model erasing all distinctions between gender, or at
least erasing them well enough to be morally acceptable, can be mapped
to many plausible but different technical definitions.  It has been
shown that these different definitions can encode different and
sometimes even conflicting moral judgments about the correct treatment
of people.  The choice between options is therefore preferably made by
multi-stakeholder debate and consensus building.  This in turn
requires that more work is done on definitions which are not only
technically feasible to implement, but can also be explained to
non-technical stakeholders.

\subsection{Technical possibilities and fashions}

Demands that an AI uses a specifically limited or imperfect world
model can often be met by using GOFAI techniques to build the model by
hand.  When we turn to the automatic building of specifically
imperfect world models, the situation gets more complicated.  The
currently most fashionable branches of machine learning all produce
opaque, black-box world models which cannot easily be edited to
include specific imperfections.  Deep learning and model-free
reinforcement learning both produce world models which will encode the
learned knowledge that `the floor is made of wood' across a massive
set of opaque numbers.  One can imagine a software component that will
automatically edit these numbers to reliably turn the floor into lava.
But it is still an open research problem to create that software
component for these particular models.

There have been some recent developments towards resolving this open
research problem, often by routing around it.  Kusner et al.\
\cite{kusner2017counterfactual} have defined a 
criterion called {\it counterfactual fairness}, which may be used for
example to define if a learned world model is making an inappropriate
distinction between genders.  The learned world model has to be a {\it
causal world model}, a model which encodes learned knowledge into a
Pearl causal graph
\cite{pearlcausality}.  Kusner et al.\ show how a learned causal
model which unfairly makes a distinction between genders can be edited
automatically to make it fair on gender, at least according to the
criterion defined.  This work has boosted the interest in the ML
community to further improve the type of machine learning that creates
causal models.

In a recent parallel development, Pearl causal graphs are also being
used in technical work on long-term AI alignment.  Carey et al.\
\cite{carey2020incentives} use them to define crisp mathematical
criteria which can be applied to long-term alignment problems for
autonomous reward-maximizing AIs.  Holtman
\cite{holtman2021cf} shows how one can bypass the
problem of having to edit a complete and correct black-box world
model, by instead wrapping the model into a Pearl causal graph which
will then produce specifically incorrect predictions.  They are
usefully incorrect by suppressing the emergent incentive which an
advanced AI may develop to disable its own built-in safety mechanisms.

The above modification techniques create AIs that are worse economic
actors, if we measure economic performance by their reward function
alone. But they make them into better socioeconomic actors.

\section{Conclusions}

The main aim of this paper has been to equip participants in the AI
alignment debate with additional tools and insights that can be used to
overcome barriers.  We have considered three barriers in particular:
stakeholder uncertainty about what is technically possible, a too
narrow focus on reward maximization, and the narrative pull of current
fashions in machine learning.  The framing of cognitive architectures
can be used to overcome these barriers, and to give more debating
power to non-specialist stakeholders.

The cognitive architectures of modern governments and companies have
many features that were explicitly designed to make them more aligned
with human goals and needs.  Stakeholders in the AI alignment debate
can locate these features, and then demand that these are also
designed into the cognitive architectures of powerful AIs.  We have
reviewed three somewhat unfashionable features which might be
demanded: automated compliance officers, pro-social reward function
components, and specifically incorrect world models.  We have also
emphasized the possibility to meet such demands with GOFAI techniques.

We also considered how stakeholders should calibrate their
interactions with individual ML researchers, if they want to encourage
or fund more AI alignment research.  We do not believe it is
productive for such stakeholders to expect or demand that ML
researchers will take the lead in solving all AI alignment problems.
We prefer a framing which states that many of the open research
problems in AI alignment are broad systems engineering problems, not
ML research problems.

The idea of demanding that powerful AIs use specifically inaccurate 
world models is based in part on our earlier work in \cite{holtman2021cf}.

\subsection{Related work and further reading}
\label{comparerelwork}

We now discuss some connections and related work not already mentioned
elsewhere in this paper.  Our description of reinforcement learning in
section 5 emphasizes the building and use of predictive world models.
The same emphasis is present in the planning-as-inference model of
human cognition developed by Botvinick and Toussaint
\cite{botvinick2012planning}.

Gabriel \cite{gabriel2020artificial} conducts a detailed examination
of the questions of moral philosophy that arise in the alignment
context, to develop propositions on what AI alignment research and
debate should be about.  There are close parallels between these
propositions and the approach taken in this paper.  Gabriel speculates
that the methods we use to build AI may influence the kind of values
we are able to encode.  This paper supports and illustrates this thesis.
We have shown how certain types of moral demands might best be
supported by introducing technical elements unrelated to reward
maximization and reward function design.

Greene et al.\
\cite{Greene2019BetterNC} and Mittelstadt
\cite{mittelstadt2019principles} discuss the narrative pull of
certain conventional framings around business and professional ethics,
and how these have shaped the recent AI alignment debate.  They call
for the debate to move beyond these particular framings.

Sambasivan et al.\ \cite{sambasivan2021everyone} also examine fashion
and safety.  They discuss how among AI technologists, it is not
fashionable to work on the problem of quality assurance for training
data.  This can severely impact safety when such training data is used
to create AIs deployed in high-stakes domains.

Dotan and Milli \cite{dotan2020value} explore the rise and fall of
fashions in ML research through the lens of philosophy of science.
Sutton's blog post {\it The bitter lesson}
\cite{sutton_bitter} makes for interesting between-the-lines reading, if
one reads it as a promise of what researchers choosing a career in ML
will never be required to do.  Counterpoints to Sutton are
offered by Brooks \cite{brooks_better} and Marcus
\cite{marcus2020next}.

\subsection*{Acknowledgments}

We thank Ben Smith, Roland Pihlakas, and the anonymous reviewers for
their comments which led to improvements in this paper.  This paper
reflects a research effort conducted while the author was working
as an independent researcher.  No external funding was sought or
received.

\subsection*{Version history}

The first version of this paper was presented at the PERLS Workshop at
35th Conference on Neural Information Processing Systems (NeurIPS
2021).  This second arXiv version extends the first version by adding
four figures, and some extra lines of text to integrate the figures
into the narrative.


\bibliographystyle{plainnat}
\bibliography{refsdemanding}


\end{document}